\documentclass[journal]{IEEEtran}

\usepackage{cite}
\usepackage{array}
\usepackage{graphicx}
\usepackage{url}
\usepackage{amsfonts}
\usepackage{multirow}
\usepackage{subcaption}
\usepackage{tabularx}
\usepackage{amsmath}
\usepackage{comment}
\usepackage{array}
\usepackage{amssymb}
\usepackage{import}
\usepackage[leftcaption]{sidecap}
\usepackage{algorithmic}
\usepackage{hyperref}

\newcommand\blfootnote[1]{%
  \begingroup
  \renewcommand\thefootnote{}\footnote{#1}%
  \addtocounter{footnote}{-1}%
  \endgroup
}

\DeclareRobustCommand*{\IEEEauthorrefmark}[1]{%
  \raisebox{0pt}[0pt][0pt]{\textsuperscript{\footnotesize #1}}%
}

\begin{document}
\title{Stereo Correspondence and Reconstruction of Endoscopic Data Challenge}
\author{
    \IEEEauthorblockN{
      M. ~Allan$^{\star}$\IEEEauthorrefmark{1}, 
      J. ~Mcleod$^{\star}$\IEEEauthorrefmark{1}, 
      C. Wang\IEEEauthorrefmark{5},
      J. C. Rosenthal\IEEEauthorrefmark{2},
      Z. Hu\IEEEauthorrefmark{2},
      N. Gard\IEEEauthorrefmark{2},
      P. Eisert\IEEEauthorrefmark{2},
      K. X. Fu\IEEEauthorrefmark{14},
      T. Zeffiro\IEEEauthorrefmark{6},
      W. Xia\IEEEauthorrefmark{3},
      Z. Zhu\IEEEauthorrefmark{7},
      H. Luo\IEEEauthorrefmark{8},
      X. Zhang\IEEEauthorrefmark{9},
      X. Li\IEEEauthorrefmark{8},
      F. Jia\IEEEauthorrefmark{8},
      L. Sharan\IEEEauthorrefmark{10},
      T. Kurmann\IEEEauthorrefmark{15},
      S. Schmid\IEEEauthorrefmark{15},
      R. Sznitman\IEEEauthorrefmark{15},
      D. Psychogyios\IEEEauthorrefmark{13},
      M. ~Azizian\IEEEauthorrefmark{1},
      D. ~Stoyanov\IEEEauthorrefmark{13},\IEEEauthorrefmark{4}
      L. ~Maier-Hein\IEEEauthorrefmark{11},
      S. ~Speidel\IEEEauthorrefmark{12}\\
    }
    \IEEEauthorblockA{
      \IEEEauthorrefmark{1}Intuitive Inc.,
      \IEEEauthorrefmark{2}Fraunhofer Institute for Telecommunications, Heinrich Hertz Institute (HHI),
      \IEEEauthorrefmark{3}University of Western Ontario,
      \IEEEauthorrefmark{4}Digital Surgery Ltd.,
      \IEEEauthorrefmark{5}Norwegian University of Science and Technology, 
      \IEEEauthorrefmark{6}Rediminds Inc.,
      \IEEEauthorrefmark{7}Harbin Institute of Technology, 
      \IEEEauthorrefmark{8}Shenzhen Institutes of Advanced Technology, 
      \IEEEauthorrefmark{9}Johns Hopkins University, 
      \IEEEauthorrefmark{10}Heidelberg University,
      \IEEEauthorrefmark{11}German Cancer Research Center (DKFZ),
      \IEEEauthorrefmark{12}National Center for Tumor Diseases (NCT),
      \IEEEauthorrefmark{13}Wellcome/EPSRC Centre for Interventional and Surgical Sciences (WEISS) UCL,
      \IEEEauthorrefmark{14}Unaffiliated,
      \IEEEauthorrefmark{15}University of Bern,
  }
}
\maketitle

\begin{abstract}

The stereo correspondence and reconstruction of endoscopic data sub-challenge was organized during the Endovis challenge at MICCAI 2019 in Shenzhen, China. The task was to perform dense depth estimation using 7 training datasets and 2 test sets of structured light data captured using porcine cadavers. These were provided by a team at Intuitive Surgical. 10 teams participated in the challenge day. This paper contains 3 additional methods which were submitted after the challenge finished as well as a supplemental section from these teams on issues they found with the dataset. 
\end{abstract}

\blfootnote{$^{\star}$ denotes equal contribution}

\section{Introduction}


Reconstruction of the surgical scene is an important problem in computer assisted surgery (CAS). It is a fundamental part of SLAM, which is a requirement for augmented reality (AR) \cite{haouchine_image_2013}, as well as a key building block of perception in automation and safety systems \cite{shademan_supervised_2016} and has even found use in diagnostic tools \cite{mahmood_polyp_2019}.

There are several methods of estimating depth in endoscopic images. The most commonly used is depth from stereo, whereby pixels are matched between the two views of a calibrated stereo camera and triangulated to provide depth measurements. The matching can be achieved through classical methods such as semi-global block matching or more recently with deep learning \cite{ye_selfsupervised_2017}. These methods are popular because they require no modification of the scene and the data required for their use can be captured with a simple recording system attaching to the endoscope. When a high quality calibration is obtained they can provide accurate results however requiring clinical teams to calibrate scopes is not a practical solution at scale and calibrations themselves can be invalidated by changing parameters such as focal lengths during the procedure \cite{pratt_practical_2014}. Calibration can be avoided by using only a single monocular view, whereby structural cues are inferred from the contents of a single image. Solutions in this space are almost entirely based on machine learning but promising results have been demonstrated in sinus endoscopy \cite{liu_self_2018}. Using structured light projectors within the endoscope has been demonstrated \cite{lin_endoscopic_2017} however obtaining reconstruction at the camera frame rate is challenging and requires specialized processing hardware. 

\begin{figure}
\captionsetup[subfigure]{labelformat=empty}
\begin{subfigure}[c]{0.24\textwidth}
\includegraphics[width=\textwidth]{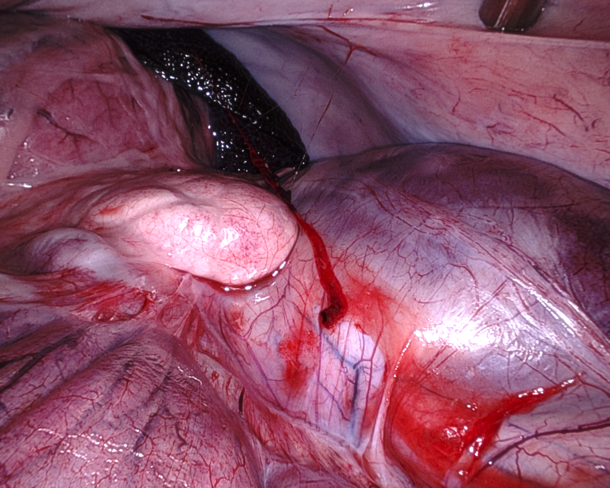}
\end{subfigure}
\hfill
\begin{subfigure}[c]{0.24\textwidth}
\includegraphics[width=\textwidth]{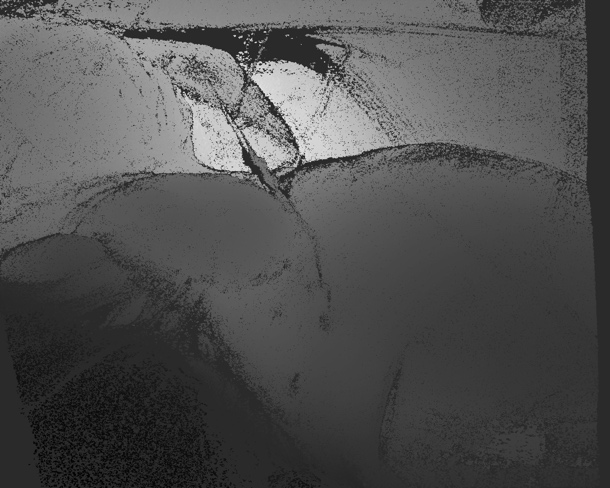}
\end{subfigure}
\caption{\label{fig:example}An example image captured by the endoscope and the corresponding depth map.}
\end{figure}


The development and evaluation of methods to accurately estimate depth in surgical images is heavily dependent on the availability of high quality datasets. The mainstream computer vision community has seen great success in algorithm development following the release of several Middlebury Stereo datasets between 2001 and 2014 \cite{scharstein_high_2003} as well as more specialized datasets from autonomous driving \cite{sun_scalability_2019, geiger_vision_2013}. Few datasets of this type exist in the surgical domain and are typically much smaller or focus on phantom or non-realistic environments \cite{stoyanov_real_2010} largely due to challenges with obtaining ground truth data. 


The stereo correspondence and reconstruction of endoscopic data (SCARED) sub-challenge was organized to help address this gap by creating a high quality dataset using porcine cadavers and a structured light projection system.

\section{Data}

The SCARED dataset consists of 7 training datasets and 2 test datasets captured using a da Vinci Xi surgical robot. Each dataset corresponds to a single porcine subject and contains between 4 and 5 keyframes. A keyframe is a single unique view of the scene where the a structured light pattern was projected into the field of view of the camera and a dense stereo reconstruction was computed \cite{scharstein_high_2003}. As the endoscope was moved using the da Vinci Xi to observe a new keypoint, the camera's pose relative to the fixed robot base was recorded using the robot's internal measurements of the position of each joint in its arms. As the scene remained static during these sequences, it was possible to obtain approximate depth maps for intermediate frames by transforming the vertices from the original keypoint and reprojecting them into the new view.

\begin{figure*}
\captionsetup[subfigure]{labelformat=empty}
\begin{subfigure}[c]{0.3\textwidth}
\includegraphics[width=\textwidth]{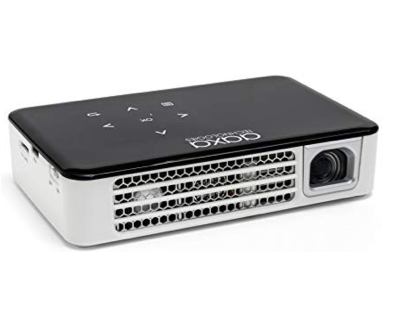}
\caption{(a)}
\end{subfigure}
\hfill
\begin{subfigure}[c]{0.3\textwidth}
\includegraphics[width=\textwidth]{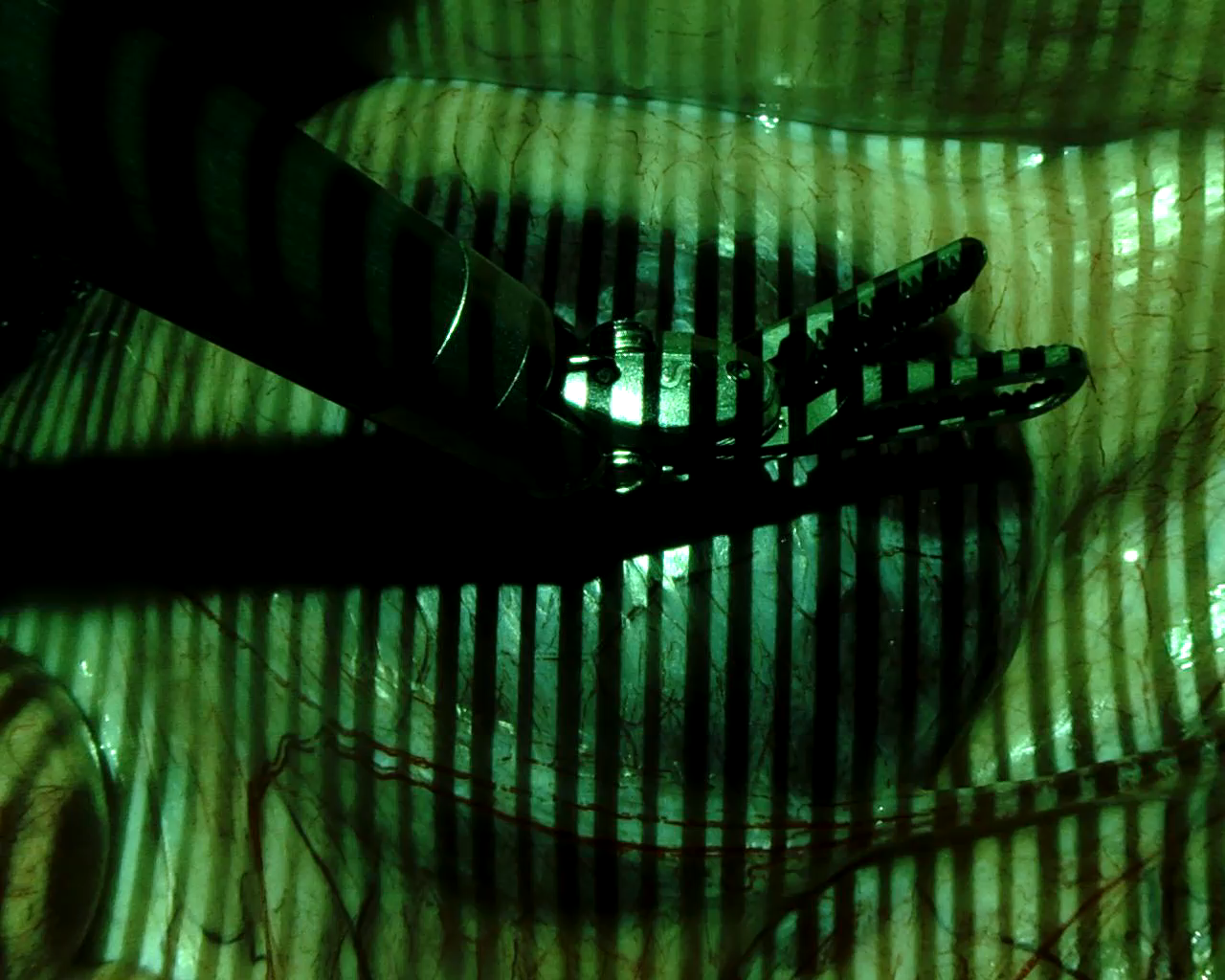}
\caption{(b)}
\end{subfigure}
\hfill
\begin{subfigure}[c]{0.3\textwidth}
\includegraphics[width=\textwidth]{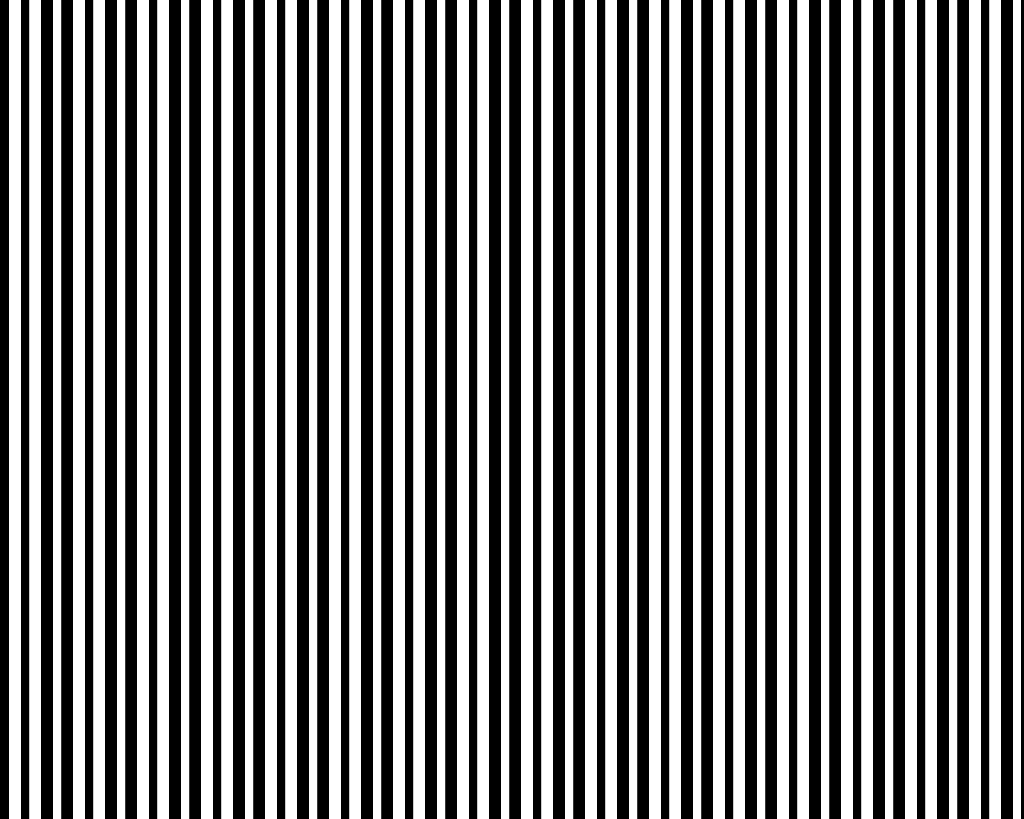}
\caption{(c)}
\end{subfigure}
\caption{\label{fig:gray_code_pattern}(a) The projector used in this data collection. (b) The view of the endoscope while the pattern is being projected. (c) A single example of a Gray code pattern used to create the dataset.}
\end{figure*}

To create the depth maps at each keypoint, the white light image of the stereo camera was first captured. Then the illuminator was turned off, and a projector was used to project a series of patterns onto the scene. The scene remained totally still during this period. The illuminator used was a AAXA P300 Neo Pico projector which provides a small form factor of $15\times3\times9$ cm and a $1280\times720$ pattern. The patterns projected were 10 bit Gray code patterns which provided a unique sequence for each pixel in the image using the method provided in \cite{scharstein_high_2003} (see Fig. \ref{fig:gray_code_pattern}). Due to challenges with positioning the projector as well, multiple projector positions had to be used to cover the entire scene. With a unique encoding for each pixel, stereo matching is straightforward and triangulation can be used to recover the scene depth.

\section{Participating Methods}

The 10 methods described in this section were submitted during the challenge time window and were the only methods considered when deciding the challenge winner.

\subsection{Fraunhofer Institute for Telecommunications, Heinrich Hertz Institute}

The submission from The Fraunhofer Institute for Telecommunications, Heinrich Hertz Institute in Germany was from Jean-Claude Rosenthal, Zhenglei Hu, Niklas Gard and Peter Eisert. This stereo pipeline consists of six steps whereas steps 2-4 are developed at Fraunhofer HHI. (1) Histogram equalization: They apply equalization using OpenCV implementation of CLAHE \cite{zuiderveld_contrast_1994} as a pre-processing step. (2) Stereo geometry analysis: They detect and match sparse feature point correspondences based on binary feature point descriptors using the hamming distance. From matched feature points they compute stereo alignment errors for 5 parameters as indicated. \cite{zilly_semantic_2011}. (3) Rectification: They correct stereo alignment errors using the matched feature points by computing a linearized F-matrix and apply a homography to rectify stereo images. They warp the right image w.r.t. to the left view \cite{zilly_joint_2010}. Remaining alignment errors are $\pm$0.25 pixels. (4) Disparity estimation: They perform a quasi-dense disparity estimation using a spatio-temporal method called patch recursive matching (patchRM) \cite{Waizenegger2016Realtime3B}. Disparity estimation fully works in parallel on a graphic card with CUDA support. The procedure makes use of a statistical approach on sub-pixel level to estimate new correspondences. These correspondences are distributed into the local neighborhood where new correspondences are determined within the next iteration. This independent propagation of new estimated correspondences guarantees that the whole disparity map is constantly updated locally while propagating the results into the spatial-temporal consistent global map. (5) 3D reconstruction and depth filtering: They generate an almost complete raw depth z-map from step 4 which uses a depth range histogram to remove outliers and mismatches during this process. Therefore, they make use of pre-known ground truth data from training sets to derive a heuristic with reasonable endoscopic depth ranges using (a) the smallest “near range” 14.676 mm (D3k5) and (b) the largest “far range” is 245.554mm (D6k5) as working distances. (6) Depth Completion: They complete the final depth map as some depth values have been marked invalid in step 4 and 5. Therefore, they use a deep neural network. It is a 2-step process which is independent of the RGBD/depth sensor. First, object occlusion boundaries and surface normals are estimated using a CNN network in the RGB image. Second, in a global surface optimization step depth values are used as a soft constraint to solve for missing depths near estimated object boundaries using the surface normals \cite{zhang2018deep}. Remarks: The pipeline has a short initialization phase of 15-25 frames for step 2 and 3. Afterwards, the estimated homography matrix to rectify the stereo images is considered static for the rest of the sequence. Therefore, these steps are no longer executed except the warping of the right image w.r.t. to the left view. 

\begin{table}
    \centering
    \small
    \begin{tabular}{ 
        p{0.17\textwidth} | 
        >{\centering}p{0.1\textwidth} | 
        >{\centering}p{0.1\textwidth} | 
        >{\centering}p{0.1\textwidth} | 
        >{\centering}p{0.1\textwidth} | 
        >{\centering}p{0.1\textwidth} | 
        p{0.05\textwidth}
    }
      & Keyframe 1 & Keyframe 2 & Keyframe 3 & Keyframe 4 & Keyframe 5 & Average \\
    \hline
    Congcong Wang   & 6.30 & \textbf{2.15} & 3.41 & 3.86 & 4.80 & 4.10 \\
    J.C. Rosenthal  & 8.25 & 3.36 & 2.21 & \textbf{2.03} & \textbf{1.33} & \textbf{3.44} \\
    KeXue Fu        & 30.49 & 18.32 & 19.73 & 19.30 & 16.86 & 20.94 \\
    Trevor Zeffiro  & 7.91 & 2.97 & \textbf{1.71} & 2.52 & 2.91 & 3.60  \\
    Wenyao Xia      & \textbf{5.70} & 7.18 & 6.98 & 8.66 & 5.13 & 6.73 \\
    Zhu Zhanshi     & 14.64 & 7.77 & 7.03 & 7.36 & 11.22 & 9.60  \\
    Huoling Luo     & 29.68 & 16.36 & 13.71 & 22.42 & 15.43 & 19.52 \\
    Xiran Zhang     & 12.53 & 6.13 & 3.60 & 3.34 & 5.07 & 6.13 \\
    Xiaohong Li     & 34.42 & 20.66 & 17.84 & 27.92 & 13.00 & 22.77 \\
    Lalith Sharan   & 30.63 & 46.51 & 45.79 & 38.99 & 53.23 & 43.03 \\
    \hline
    Dimitris Psychogyios 1 & 7.73 & 2.07 & 1.94 & 2.63 & 0.62 & 3.00 \\
    Dimitris Psychogyios 2 & 7.41 & 2.03 & 1.92 & 2.75 & 0.65 & 2.95 \\
    Sebastian Schmid & 7.61 & 2.41 & 1.84 & 2.48 & 0.99 & 3.07 \\
    \end{tabular}
    \caption{\label{tab:dataset_1_results} The mean absolute depth error in mm for test dataset 1. The best method submitted during the challenge period is shown in bold.}
\end{table}
    
\subsection{University of Western Ontario}

The submission from the Robarts Institute at the University of Western Ontario, Canada was from Wenyao Xia. This method used multi-scale cost volume filtering with mean-squared-error loss wtth post-processing with guided median filter. A regression model was also trained to correct rectification errors.

\subsection{Norwegian University of Science and Technology}

The submission from the Norwegian University of Science and Technology was from Congcong Wang. This method used a variational disparity estimation method on the coarsest level of a multiscale pyramid \cite{wang2018liver}. The disparity map was upsampled using a modified bilateral filter. 

\subsection{Rediminds Inc.}

The submission from the Rediminds Inc. USA was from Trevor Zeffiro. This method trained a Pyramid Stereo Matching network (PSMNet) using an Gaussian based interpolation strategy to remove holes in the training data.

\subsection{Unknown}

The unaffiliated submission is from KeXue Fu. This method uses unsupervised training of a DNN to predict the disparity and then reproject using the stereo calibration parameters.

\subsection{Harbin Institute of Technology}

The submission from the Harbin Institute of Technology, China was from Zhu Zhanshi. This method used a 6 level UNet with multi-scale loss on appearance matching, disparity smoothness and left-right consistency.

\subsection{Shenzhen Institute for Advanced Technology (1)}

The first submission from the Shenzhen Institute for Advanced Technology, China was from Huoling Luo and Fucang Jia. This method used a modified VGG encoder with skip connections to a dual branch decoder to directly predict a disparity map. 4 neighborhood smoothing is used as post-processing. 

\subsection{Johns Hopkins University}

The submission from the Johns Hopkins University, USA was from Xiran Zhang. This method used the Pyramid Stereo Matching network (PSMNet), discarding inaccurate pixels in the training data using SGBM as a filter and ICP to generate a new point cloud which is reprojected to the create high quality training images. L1 loss is used to regularize the results. 

\subsection{Shenzhen Institute for Advanced Technology (2)}

The second submission from the Shenzhen Institute for Advanced Technology, China was from Xiaohong Li and Fucang Jia. This method used a CycleGAN to synthesize a depth map in a 2 stage process directly from a rectified stereo pair.

\subsection{Heidelberg University}

The submission from the Heidelberg University, Germany was from Lalith Sharan. This method used a UNet trained to minimize a mean absolute error loss with a smoothness regularization.

\section{Post-Challenge Methods}

The 3 methods described in this section were submitted in the months after the challenge finished so were not considered in the final challenge ranking. 

\subsection{Wellcome/EPSRC Centre for Interventional and Surgical Sciences (WEISS) UCL (1)}
\label{ssect:deep_pruner}
The first submission from UCL is from Dimitris Psychogyios. This method used Deep Pruner \cite{duggal2019deeppruner}. The model was pretrained on the Sceneflow dataset \cite{mayer2016large}. Outlier points were manually removed from ground truth and stereo frames were rectified using OpenCV. Disparity images were generated in the left rectified frame of reference. During training, a data augmentation scheme consisted of random crops of size 256x512 and color normalization was used. They ignored the interpolation sequences as well as datasets 4 and 5 due to high calibration errors. This left 25 samples, of which they allocated 18 for training and 7 for evaluation. Finally, inferred disparities were reconstructed in 3D using the provided calibration parameters. Training was performed using a batch size of 1 for 290 epochs. The Adam optimizer \cite{kingma2014adam} with a learning rate of $10^{- 4}$ , $\beta_{1} = 0.9$ and $\beta_{2} = 0.999$. 

\subsection{Wellcome/EPSRC Centre for Interventional and Surgical Sciences (WEISS) UCL (2)}

The  second  submission  from  UCL  is  also  from  Dimitris Psychogyios. This method makes use of Hierarchical deep Stereo Matching  network  (HSM)\cite{yang2019hierarchical}. It  followed  the  same  process as illustrated in Subsection IV-A with the exception of a batchsize of 2 and 350 epochs of training.

\subsection{University of Bern}

The third late submission was from Sebastian Schmid and Tom Kurmann at University of Bern. Similar to previous methods such as pwc-net \cite{pwc_net_2018} and gwc-net \cite{guo2019group} they base their method on a 3D cost volume optimization for stereo matching. They extend the previous method by using two parallel pipelines to predict both left and right disparities during training. This allows them to incorporate photometric losses which enforce a correct reconstruction of the right image with the left image and disparities and vice-versa. They optimize the computationally expensive 3D cost volume and subsequent 3D convolution network by quantizing the cost volume. Rather than using a disparity stride of 1 when generating the volume, they use a dynamic stride which is based on the distribution of the ground truth disparities. They use k-means clustering of the ground truth disparities to extract n-centroids, which are used as the disparities in the cost volume. Without a loss of accuracy, they can reduce the number of disparities by 40\%, resulting in a decrease in inference time of 40\% (79.1ms). Furthermore, this reduces the memory footprint, allowing for larger batch sizes or higher resolution during training. Their loss function is the weighted sum of the L1-loss of the prediction and ground truth, the photometric loss and a disparity smoothness term.

\section{Results}

Evaluation is performed on 2 test datasets, each containing 5 keyframes. Similarly to the training data, keyframes 1-4 all have interpolation sequences from keyframe $N$ to keyframe $N+1$ and keyframe 5 is a single image. The metric we use is the mean absolute error in mm of the depth measurement at each pixel. During the interpolation sequences, we mask the pixels which do not have associated ground truth so they are not considered in the error measurement and additionally we discard frames for which less than 10\% of the frames have ground truth measurements. 

\subsection{Test Dataset 1}

\begin{figure*}
\captionsetup[subfigure]{labelformat=empty}
\begin{subfigure}[b]{0.5\textwidth}
\includegraphics[width=\textwidth]{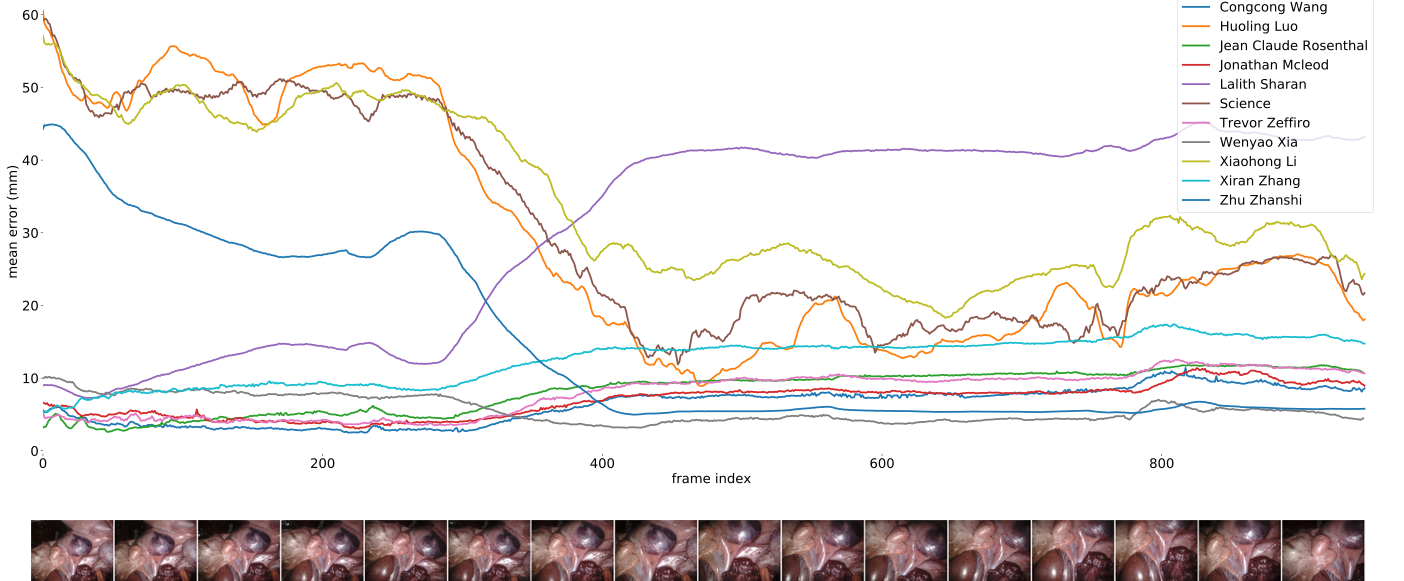}
\caption{Keyframe 1}
\end{subfigure}
\hfill
\begin{subfigure}[b]{0.5\textwidth}
\includegraphics[width=\textwidth]{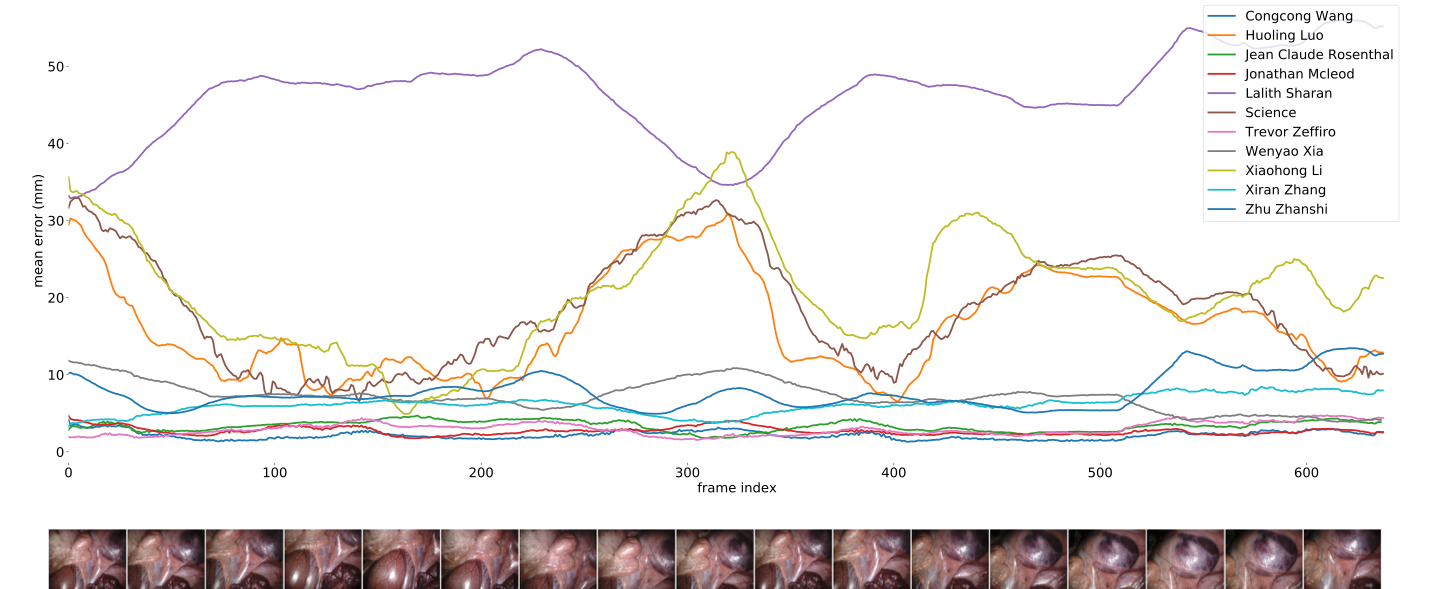}
\caption{Keyframe 2}
\end{subfigure}
\caption{\label{fig:test_dataset_1_keyframe_1_2}Mean absolute error plots for each frame of test dataset 1, keyframes 1 and 2. }
\end{figure*}

\begin{figure*}
\captionsetup[subfigure]{labelformat=empty}
\begin{subfigure}[b]{0.5\textwidth}
\includegraphics[width=\textwidth]{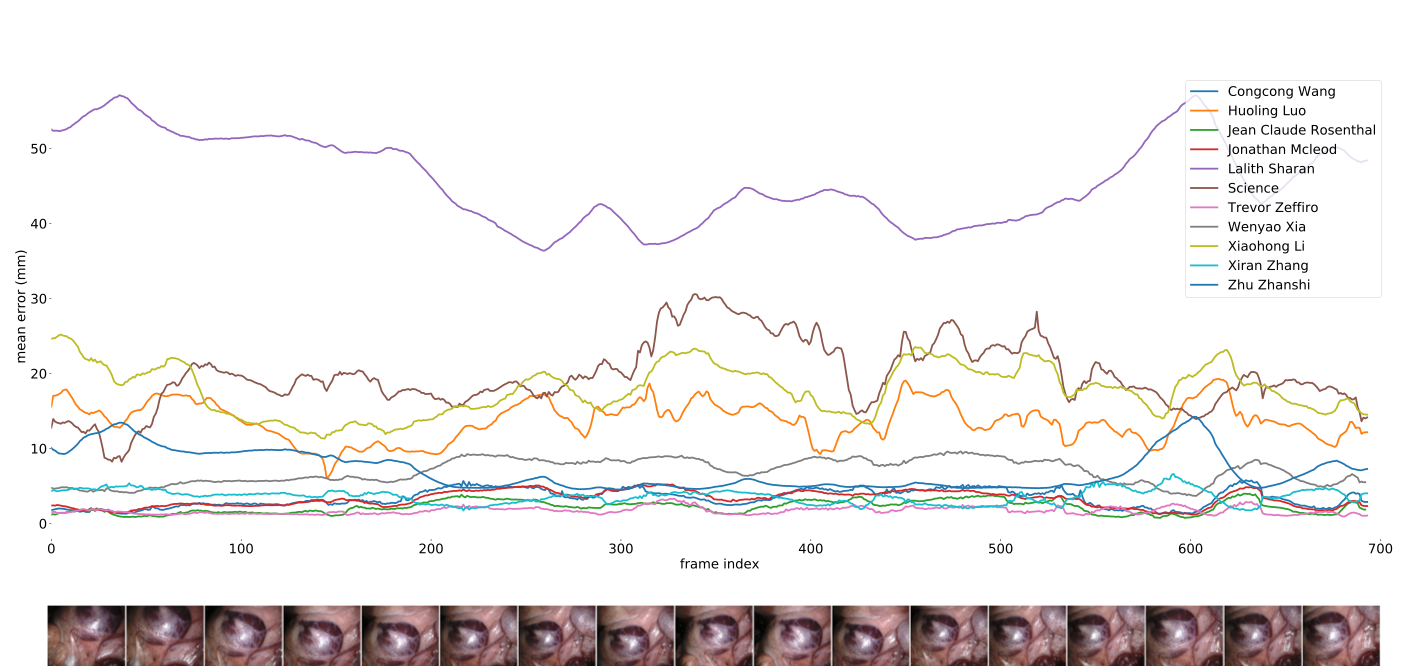}
\caption{Keyframe 3}
\end{subfigure}
\hfill
\begin{subfigure}[b]{0.5\textwidth}
\includegraphics[width=\textwidth]{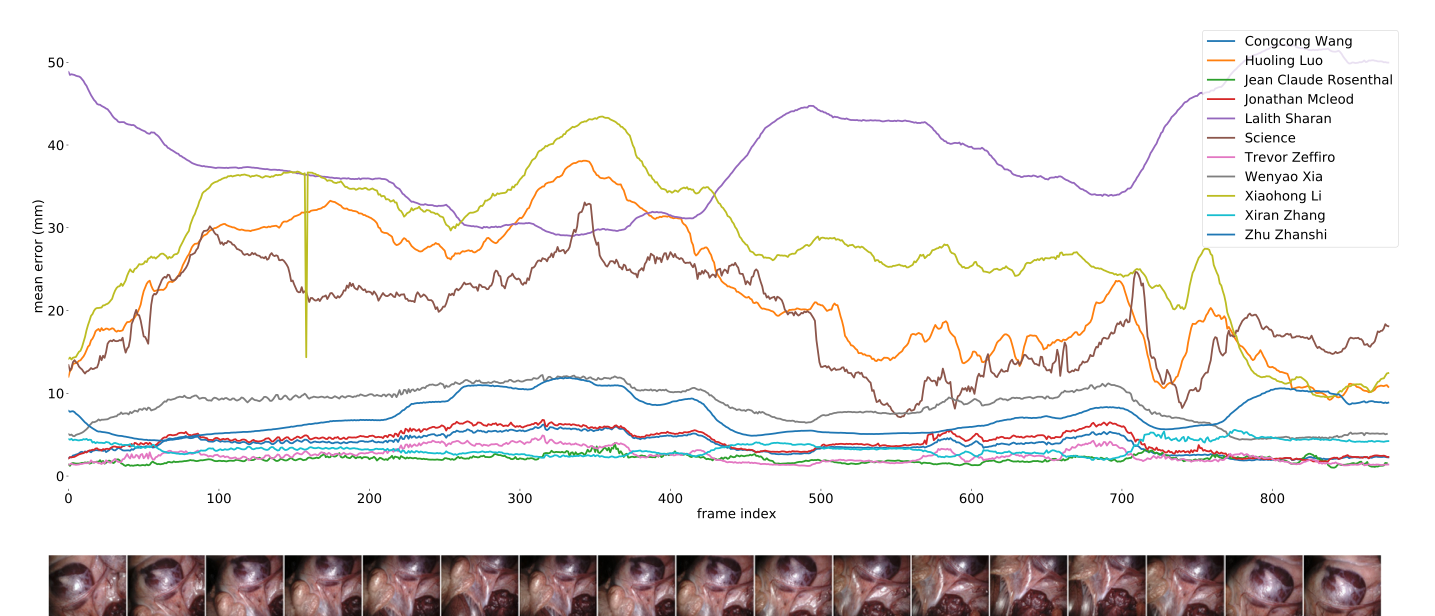}
\caption{Keyframe 4}
\end{subfigure}
\caption{\label{fig:test_dataset_1_keyframe_3_4}Mean per-pixel error plots for each frame of test dataset 1, keyframes 3 and 4.}
\end{figure*}

The overall numerical errors on test dataset 1 from each method are shown in Fig. \ref{tab:dataset_1_results} and the plots of error per frame for keyframes with motion (1-4) are shown in Fig. \ref{fig:test_dataset_1_keyframe_1_2} and Fig. \ref{fig:test_dataset_1_keyframe_3_4}. 
\begin{table*}
\centering
\small
\begin{tabular}{ 
    p{0.17\textwidth} | 
    >{\centering}p{0.1\textwidth} | 
    >{\centering}p{0.1\textwidth} | 
    >{\centering}p{0.1\textwidth} | 
    >{\centering}p{0.1\textwidth} | 
    >{\centering}p{0.1\textwidth} | 
    p{0.05\textwidth}
}
  & Keyframe 1 & Keyframe 2 & Keyframe 3 & Keyframe 4 & Keyframe 5 & Average \\
\hline
Congcong Wang   & 6.30 & \textbf{2.15} & 3.41 & 3.86 & 4.80 & 4.10 \\
J.C. Rosenthal  & 8.25 & 3.36 & 2.21 & \textbf{2.03} & \textbf{1.33} & \textbf{3.44} \\
KeXue Fu        & 30.49 & 18.32 & 19.73 & 19.30 & 16.86 & 20.94 \\
Trevor Zeffiro  & 7.91 & 2.97 & \textbf{1.71} & 2.52 & 2.91 & 3.60  \\
Wenyao Xia      & \textbf{5.70} & 7.18 & 6.98 & 8.66 & 5.13 & 6.73 \\
Zhu Zhanshi     & 14.64 & 7.77 & 7.03 & 7.36 & 11.22 & 9.60  \\
Huoling Luo     & 29.68 & 16.36 & 13.71 & 22.42 & 15.43 & 19.52 \\
Xiran Zhang     & 12.53 & 6.13 & 3.60 & 3.34 & 5.07 & 6.13 \\
Xiaohong Li     & 34.42 & 20.66 & 17.84 & 27.92 & 13.00 & 22.77 \\
Lalith Sharan   & 30.63 & 46.51 & 45.79 & 38.99 & 53.23 & 43.03 \\
\hline
Dimitris Psychogyios 1 & 7.73 & 2.07 & 1.94 & 2.63 & 0.62 & 3.00 \\
Dimitris Psychogyios 2 & 7.41 & 2.03 & 1.92 & 2.75 & 0.65 & 2.95 \\
Sebastian Schmid & 7.61 & 2.41 & 1.84 & 2.48 & 0.99 & 3.07 \\
\end{tabular}
\caption{\label{tab:dataset_1_results} The mean absolute depth error in mm for test dataset 1. The best method submitted during the challenge period is shown in bold.}
\end{table*}

\subsection{Test Dataset 2}

\begin{figure*}
\captionsetup[subfigure]{labelformat=empty}
\begin{subfigure}[b]{0.5\textwidth}
\includegraphics[width=\textwidth]{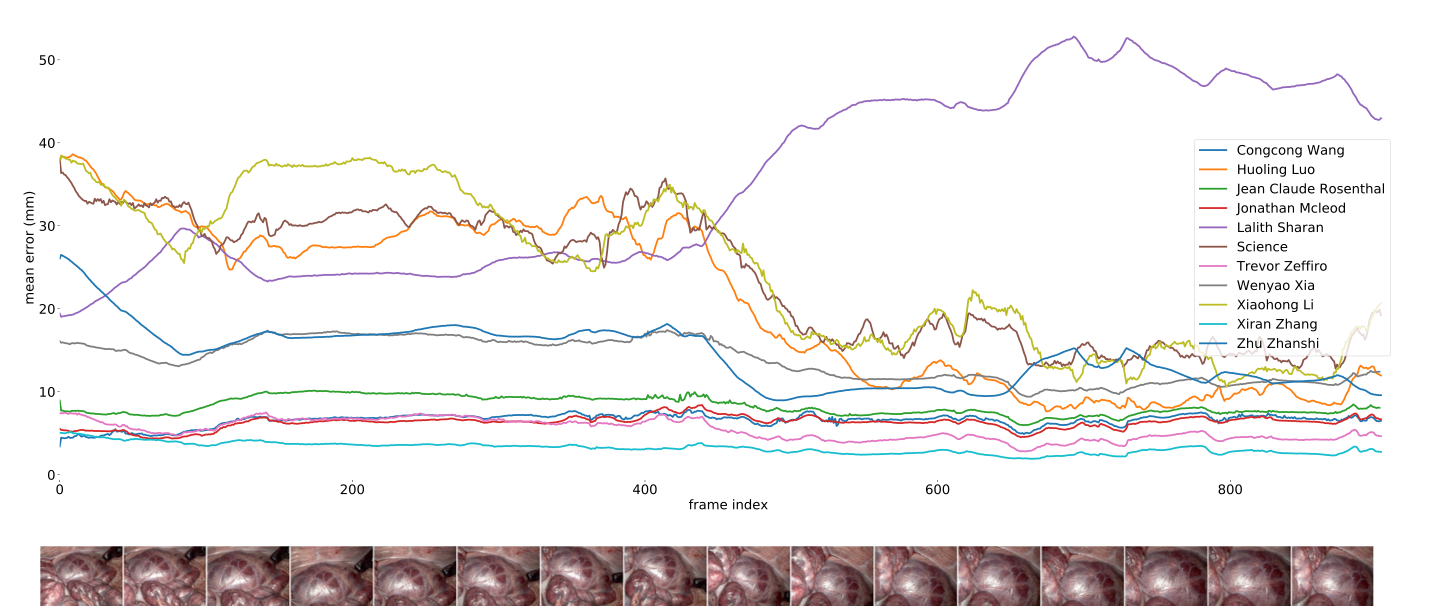}
\caption{Keyframe 1}
\end{subfigure}
\hfill
\begin{subfigure}[b]{0.5\textwidth}
\includegraphics[width=\textwidth]{figures/results/dataset_2_keyframe_1.png}
\caption{Keyframe 2}
\end{subfigure}
\caption{\label{fig:test_dataset_2_keyframe_1_2}Mean per-pixel error plots for each frame of test dataset 2, keyframes 1 and 2. }
\end{figure*}

\begin{figure*}
\captionsetup[subfigure]{labelformat=empty}
\begin{subfigure}[b]{0.5\textwidth}
\includegraphics[width=\textwidth]{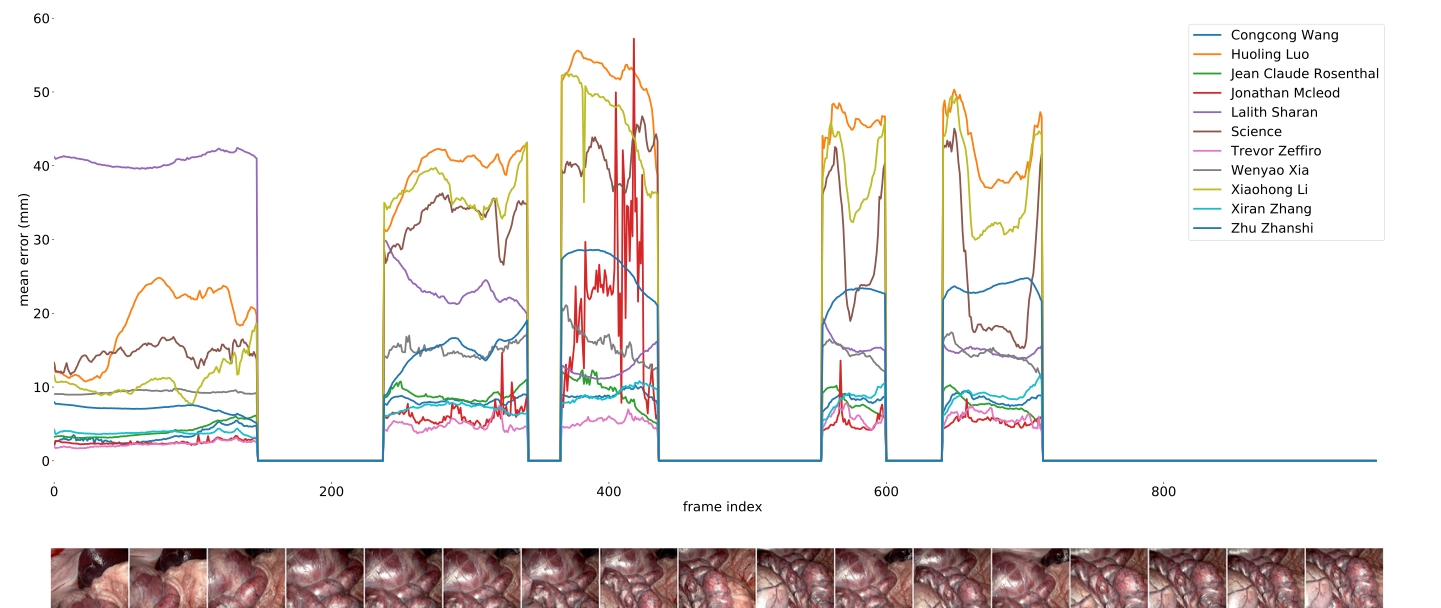}
\caption{Keyframe 3}
\end{subfigure}
\hfill
\begin{subfigure}[b]{0.5\textwidth}
\includegraphics[width=\textwidth]{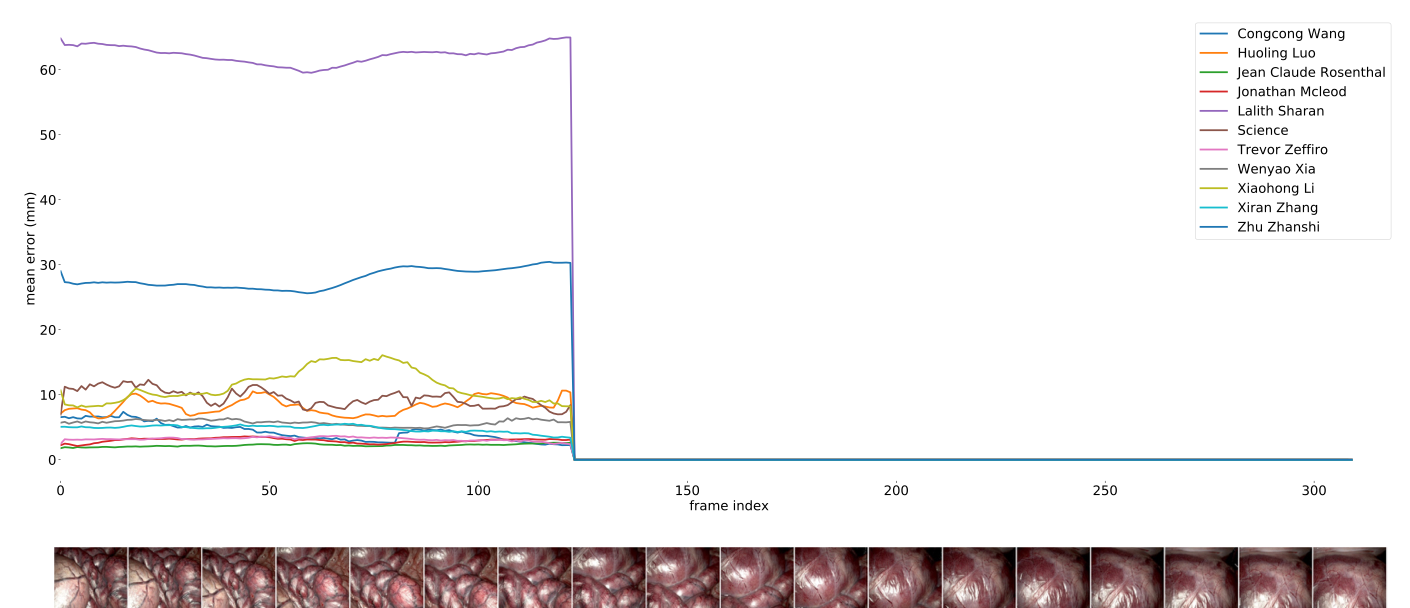}
\caption{Keyframe 4}
\end{subfigure}
\caption{\label{fig:test_dataset_2_keyframe_3_4}Mean per-pixel error plots for each frame of test dataset 2, keyframes 3 and 4. }
\end{figure*}

The overall numerical errors on test dataset 2 from each method are shown in Fig. \ref{tab:dataset_2_results} and the plots of error per frame for keyframes with motion (1-4) are shown in Fig. \ref{fig:test_dataset_2_keyframe_1_2} and Fig. \ref{fig:test_dataset_2_keyframe_3_4}. 

\begin{table*}
\centering
\small
\begin{tabular}{ 
    p{0.17\textwidth} | 
    >{\centering}p{0.1\textwidth} | 
    >{\centering}p{0.1\textwidth} | 
    >{\centering}p{0.1\textwidth} | 
    >{\centering}p{0.1\textwidth} | 
    >{\centering}p{0.1\textwidth} | 
    p{0.05\textwidth}
}
  & Keyframe 1 & Keyframe 2 & Keyframe 3 & Keyframe 4 & Keyframe 5 & Average \\
\hline
Congcong Wang   & 6.57 & 2.56 & 6.72 & 4.34 & 1.19 & 4.28 \\
J.C. Rosenthal  & 8.26 & 2.29 & 7.04 & \textbf{2.22} & \textbf{0.42} & 4.05 \\
KeXue Fu        & 23.71 & 15.46 & 26.43 & 9.60 & 10.92 & 17.22 \\
Trevor Zeffiro  & 5.39 & \textbf{1.67} & \textbf{4.34} & 3.18 & 2.79 & \textbf{3.47} \\
Wenyao Xia      & 13.80 & 6.85 & 13.10 & 5.70 & 7.73 & 9.44 \\
Zhu Zhanshi     & 14.41 & 12.55 & 16.30 & 27.87 & 34.86 & 21.20 \\
Huoling Luo     & 20.83 & 11.27 & 35.74 & 8.26 & 14.97 & 18.21 \\
Xiran Zhang     & \textbf{3.20} & 3.30 & 6.75 & 4.79 & 3.91 & 4.39 \\
Xiaohong Li     & 24.58 & 16.80 & 29.92 & 11.37 & 19.93 & 20.52 \\
Lalith Sharan   & 35.46 & 50.09 & 25.24 & 62.37 & 70.45 & 48.72 \\
\hline
Dimitris Psychogyios 1 & 4.85 & 0.65 & 1.62 & 0.77 & 0.41 & 1.67 \\
Dimitris Psychogyios 2 & 4.78 & 1.19 & 3.34 & 1.82 & 0.36 & 2.30 \\
Sebastian Schmid & 4.33 & 1.10 & 3.65 & 1.69 & 0.48 & 2.25 \\
\end{tabular}
\caption{\label{tab:dataset_2_results} The mean absolute depth error in mm for test dataset 2. The best method submitted during the challenge period is shown in bold.}
\end{table*}

\subsection{Overall}

The overall winner of the challenge was the method with the lowest mean error across the two test datasets, which was Trevor Zeffiro of Rediminds Inc. Second place was awarded to Jean-Claude Rosenthal. 
\section{Supplemental Material on Dataset inaccuracies}

Inaccuracies were present in the provided datasets. Many of these inaccuracies were expected, such as calibration errors for the endoscope and synchronization issues between the video and kinematics in the warping of depth data from the initial frame to later frames. Subsequent efforts to estimate and correct for these efforts were made by late submissions to the challenge and their results are explained here for completeness. 

\subsection{Sebastian Schmid and Tom Kurmann}

The original dataset relies on the forward kinematics for forward projection of the point cloud. This limits the accuracy of the point clouds to the accuracy of the forward kinematics. These may suffer from positioning errors, noise and video synchronization issues making them a potential source of error in the training data. In order to minimize this error, we propose to compute the forward propagation of the point clouds using the imaging data. To do so, we perform a 4 step pipeline which detects keypoints in subsequent frames, matches them and then computes the camera's pose. Using the camera pose, the point cloud can then be projected into the image to obtain the disparity maps. More precisely, our pipeline is defined as follows:
\begin{description}
\item[$\bullet$] Extract SIFT features from the keyframe and a subsequent frame.
\item[$\bullet$] Match the features between subsequent and keyframe. The aim is to find as many feature pairs as possible.
\item[$\bullet$] Use the matched features to estimate the pose of the endoscope, substituting the given forward kinematics. This can be formulated as a perspective-n-point problem, where the projections of a set of $n$ 3D points onto an image plane are known and the pose of the camera has to be calculated. The pose is expressed as rotation $\mathbf{R}$ and translation $\mathbf{t}$.
\item[$\bullet$] Apply $\mathbf{R}$ and $\mathbf{t}$ to the point cloud of the keyframe to get a point cloud for the subsequent frame.
\end{description}

\subsection{Dimitris Psychogyios}

\subsubsection{Calibration Errors}

\begin{figure*}
\begin{subfigure}[b]{0.32\textwidth}
\includegraphics[width=\textwidth]{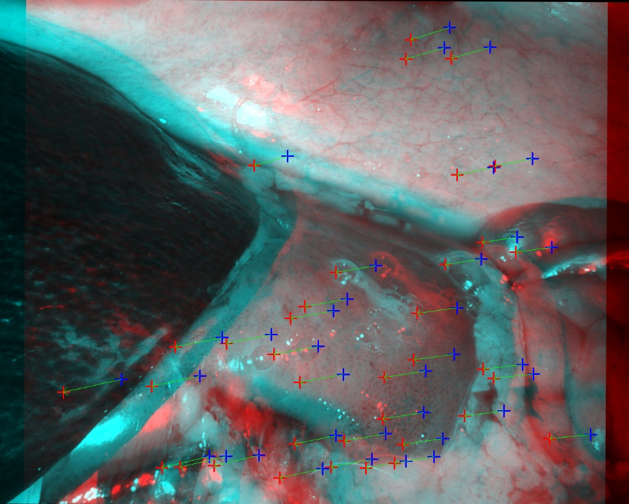}
\caption{}
\end{subfigure}
\hfill
\begin{subfigure}[b]{0.32\textwidth}
\includegraphics[width=\textwidth]{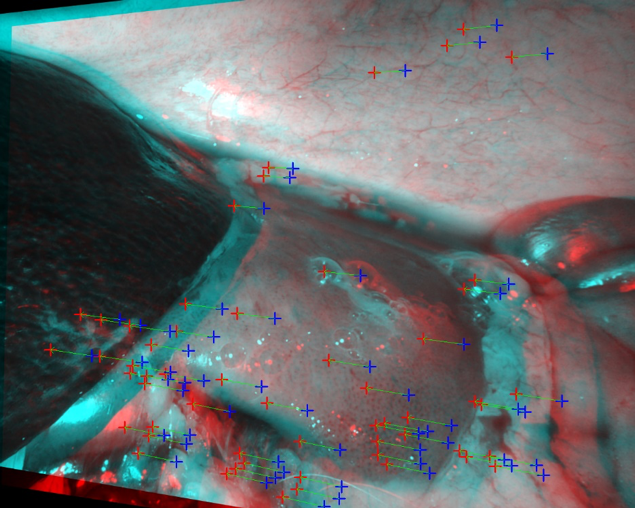}
\caption{}
\end{subfigure}
\hfill
\begin{subfigure}[b]{0.32\textwidth}
\includegraphics[width=\textwidth]{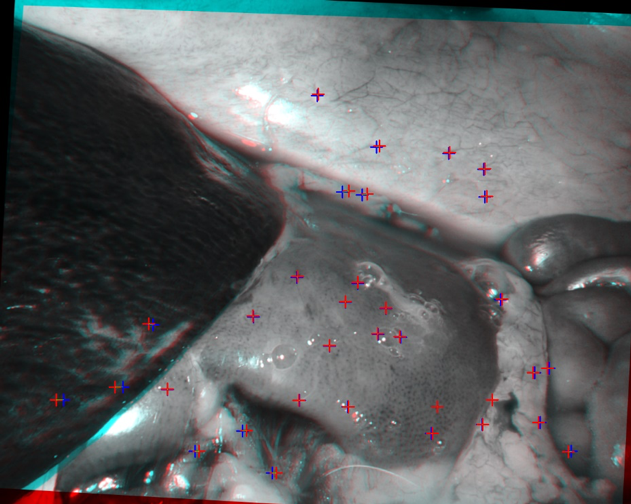}
\caption{}
\end{subfigure}
\caption{\label{fig:supplemental_calib_error}(a) Stereo anaglyph of dataset 4, keyframe 1 showing rectified frames, using the provided calibration parameters, with sparse feature matching overlay. Corresponding features do not lie on the same horizontal scanlines. (b) The same image pair where the extrinsics are estimated from feature matches. (c) Uncalibrated rectification based only on visual matches.}
\end{figure*}

The ground truth depthmap is given as a 3-channel image. Each nonzero ground truth pixel $(u,v)$ stores the 3D coordinates of a point in space which projects to $(u,v)$ pixel in the left image. Using the provided camera calibration parameters to project a 3D point, stored in ground truth’s $(u,v)$, ends up in coordinates $(u+du, v+dv)$. This effect could potentially be caused by either discretization in the depth map or calibration error. Calibration files in Datasets 4 and 5 seem to have inaccuracies. Using the provided calibration parameters to rectify frames included in those two datasets, caused large errors in the rectification as shown in Fig. \ref{fig:supplemental_calib_error}. Matches in the rectified image pairs do not end up on the same scan-lines. Estimating the Essential matrix, using Shi and Tomashi features\cite{shi1994good} and LK tracking \cite{lucas1981iterative} between the two views and using it refine the extrinsics, resulted in rectifications that had the same problem. Rectification based only on the Fundamental matrix is adequate. This leads to the conclusion that the error is mainly in the intrinsics.  

\subsubsection{Video/Kinematics Offsets}

Frames in `rgb.mp4' files and the interpolated ground truth depth images, based on robot kinematics, are not time synchronized. Overlaying the provided ground truth sequences over the corresponding videos, shows that the video is lagging. 

\subsubsection{Ground Truth-RGB misalignment}

\begin{figure*}
\captionsetup[subfigure]{labelformat=empty}
\begin{subfigure}[b]{0.48\textwidth}
\includegraphics[width=\textwidth]{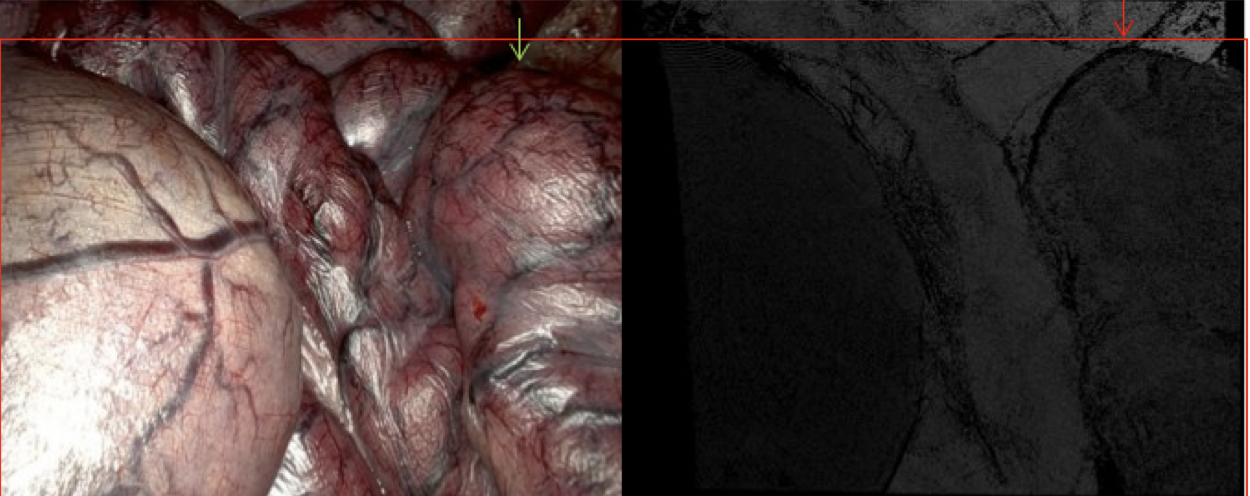}
\caption{Scanline Offset}
\end{subfigure}
\hfill
\begin{subfigure}[b]{0.48\textwidth}
\includegraphics[width=\textwidth]{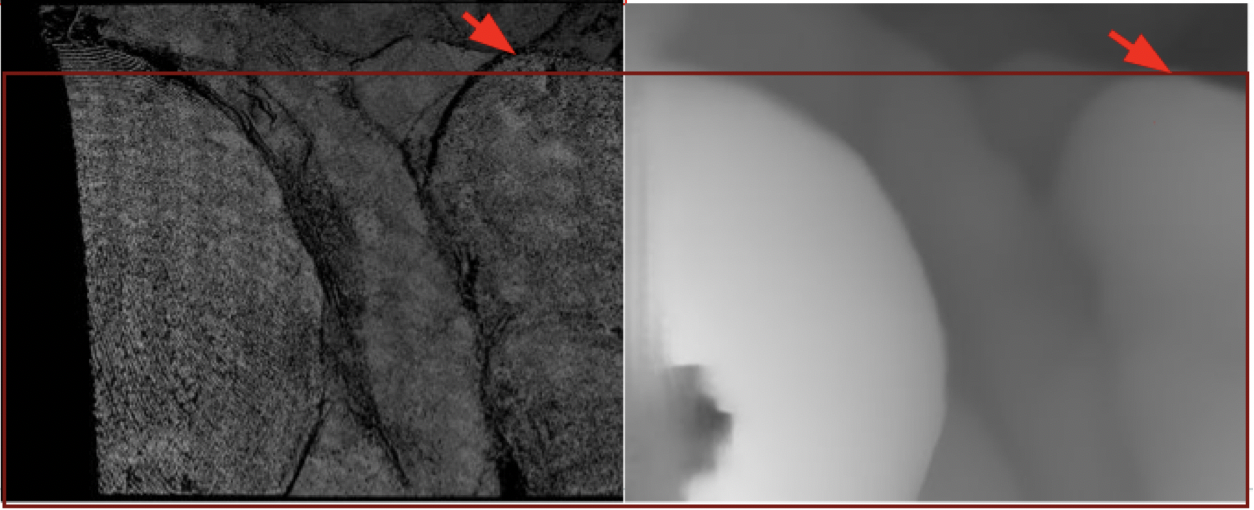}
\caption{Disparity Offset}
\end{subfigure}
\caption{\label{fig:offset_error} An example from dataset 9, keyframe 3. Ground truth depth map is shifted with respect to the RGB image. This is also visible between the disparity which is based on the ground truth data and the output of one of the networks.}
\end{figure*}

There are misalignments between the ground truth and the RGB data in datasets 8 and 9 (Figure \ref{fig:offset_error}).

\bibliographystyle{IEEEtran}
\bibliography{lib}

\end{document}